\documentclass{article}

\usepackage[nonatbib,final]{neurips_2021}
\usepackage[utf8]{inputenc} % allow utf-8 input
\usepackage[T1]{fontenc}    % use 8-bit T1 fonts
\usepackage{hyperref}       % hyperlinks
\usepackage{url}            % simple URL typesetting
\usepackage{booktabs}       % professional-quality tables
\usepackage{amsfonts}       % blackboard math symbols
\usepackage{nicefrac}       % compact symbols for 1/2, etc.
\usepackage{microtype}      % microtypography
\usepackage{xcolor}         % colors
\usepackage{wrapfig}

\usepackage{amssymb}
\usepackage{graphicx}
\usepackage{algorithm2e}
\usepackage{scalerel}
\def\msquare{\mathord{\scalerel*{\Box}{gX}}}

\title{Contrastive Reinforcement Learning of Symbolic Reasoning Domains}

\author{%
  Gabriel Poesia \\
  Stanford University\\
  \texttt{poesia@cs.stanford.edu} \\
\And
  WenXin Dong \\
  Stanford University\\
  \texttt{wxd@stanford.edu} \\
\And
  Noah Goodman \\
  Stanford University \\
  \texttt{ngoodman@stanford.edu} \\
}

\begin{document}

\maketitle

\begin{abstract}
  Abstract symbolic reasoning, as required in domains such as mathematics and logic, is a key component of human intelligence. Solvers for these domains have important applications, especially to computer-assisted education. But learning to solve symbolic problems is challenging for machine learning algorithms. Existing models either learn from human solutions or use hand-engineered features, making them expensive to apply in new domains. In this paper, we instead consider symbolic domains as simple environments where states and actions are given as unstructured text, and binary rewards indicate whether a problem is solved. This flexible setup makes it easy to specify new domains, but search and planning become challenging. We introduce four environments inspired by the Mathematics Common Core Curriculum, and observe that existing Reinforcement Learning baselines perform poorly. We then present a novel learning algorithm, Contrastive Policy Learning (ConPoLe) that explicitly optimizes the InfoNCE loss, which lower bounds the mutual information between the current state and next states that continue on a path to the solution. ConPoLe successfully solves all four domains. Moreover, problem representations learned by ConPoLe enable accurate prediction of the categories of problems in a real mathematics curriculum. Our results suggest new directions for reinforcement learning in symbolic domains, as well as applications to mathematics education.
\end{abstract}

\section{Introduction}

%In the last decade, the AI community achieved major progress
%in tasks that require accurate \emph{perception}, such as image %classification,
%speech recognition and machine translation,
%in large part due to advances in deep neural networks.
%This class of models can also perform symbolic reasoning when provided a
%proper training scheme.
%This is evidenced by the success of algorithms backed by neural networks in
%learning solvers for satisfiability, automated theorem proving and
%program synthesis when provided with appropriate training data
%-- usually either human solutions or problems sampled along with a known
%solution.
% In this work, we introduce a learning algorithm for symbolic manipulation
% domains by solely relying on interaction with an environment that emits
% binary rewards and textual representations of states and actions.

Humans posses the remarkable ability to learn how to reason in symbolic
domains, such as arithmetic, algebra, and formal logic. Our aptitude for
mathematical cognition builds on specialized neural bases but extends them radically through formal education \cite{butterworth2011neural,houde2003neural,dehaene2011number,dehaene2004arithmetic}.
Learning to reason in symbolic domains poses an important challenge for artificial intelligence research. As we describe below, this type of reasoning has unique features that distinguish it from domains in which machine learning has had recent success.

From a practical viewpoint, since symbolic reasoning skills span years of instruction in school, advances in symbolic reasoning may have a large impact on education. In particular, automated tutors equipped with step-by-step solvers can provide personalized help for students working through problems \cite{ritter2007cognitive},
and aid educators in curriculum and course design by semantically relating exercises based on their solutions  \cite{mccalla1992search,melis2004activemath}.
Indeed, studies have found automated tutors capable of yielding similar \cite{anderson1985intelligent,ma2014intelligent}
or larger \cite{vanlehn2011relative} educational gains than human tutors.
While \emph{solving} problems alone does not necessarily translate to
good \emph{teaching}, automated tutors typically have powerful domain models
as their underlying foundation.

However, even modest mathematical domains
are challenging to solve. As an example, consider
solving linear equations step-by-step using low-level axioms, such
as associativity, reflexivity and operations with constants.
This formulation allows all solution strategies
that humans employ to be expressed as combinations of few simple rules,
making it attractive for automated tutors \cite{ritter2007cognitive,o2019automatic}.
But while formulating the domain is simple, obtaining a general solver is not.
Na\"ive search is infeasible due to the combinatorial solution space.
As an example, the search-based solver used in the recent Algebra Notepad tutor \cite{o2019automatic} is limited to solutions of up to 4 steps.
An alternative is manually writing expert solver heuristics.
Again, even for a domain such as high-school algebra, this route is
difficult and error-prone.
As we describe in Section~\ref{sec:eval}, we evaluated Google MathSteps,
a library that backs educational applications with a step-by-step algebra solver, on the equations from the Cognitive Tutor Algebra \cite{ritter2007cognitive}
dataset. MathSteps only succeeded in 76\% of the test set, revealing
several edge cases in its solution strategies.
Thus, even very complex expert-written strategies may have surprising gaps.

% For example, it fails to solve $1 = -2  - 3/x$. Thus, even very complex
% expert-written search strategies may have surprising gaps.

An alternative could be to \emph{learn} solution strategies
via Reinforcement Learning (RL).
We formulate symbolic reasoning as an RL problem of deterministic environments that execute domain rules and give a binary
reward when a problem is solved.
Since we aim for generality, we assume a domain-agnostic interface with the environment:
states and actions are given to agents as unstructured text.
These domains have several idiosyncrasies that make them challenging for RL.
First, trajectories are unbounded, since axioms might always be applicable and lead to
new states (e.g. adding a constant to both sides of an equation).
Second, agents have no direct access to the underlying structure of the domain,
only observing strings and sparse binary rewards.
Finally, each problem only has one success state (e.g. \texttt{x = number}, in equations).
These properties rule out many popular algorithms for RL. 
For instance, Monte Carlo Tree Search (MCTS, \cite{chaslot2006monte}) uses random policy rollouts to train its value estimates. If the solution state is unique, such rollouts only find
non-zero reward if they happen to find the complete solution.
Thus, MCTS fails to guide search toward solutions \cite{agostinelli2019solving}.
Indeed, as we show in Section \ref{sec:eval}, Deep Q-Learning, and other algorithms
that are based on estimating expected rewards, perform poorly in these symbolic domains.

\begin{figure}
    \centering
    \includegraphics[width=\textwidth]{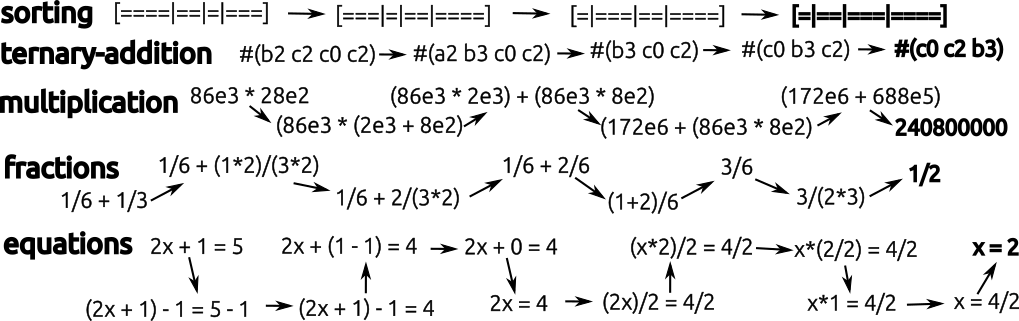}
    \caption{Example of one problem and step-by-step solution in each
    CommonCore environment.
    In \texttt{equations}, a problem is a linear equation
    with the four basic operations and arbitrary parentheses,
    and actions are applications of low-level axioms (e.g. commutativity,
    associativity, distributivity,
    calculations with constants and applying operations on both sides).
    \texttt{ternary-addition} simulates step-by-step arithmetic with
    carry in base 3: inputs are sequence of $(digit, power)$ pairs,
    where letters are digits ($\texttt{a} = 0$, $\texttt{b} = 1$,
    $\texttt{c} = 2$) and numbers are powers of 3 (so $\texttt{b2} = 1\cdot{}3^2$). Adjacent digits can be added together, and
    digits $0$ can be eliminated. In the example,
    we start with
    $1\cdot{}3^2 + 2\cdot{}3^2 + 2\cdot{}3^0 + 2\cdot{}3^2$
    and obtain $2\cdot{}3^0 + 1\cdot{}3^1 + 2\cdot{}3^2$.
    In \texttt{multiplication}, addition and single-digit
    multiplication are primitive operations, and must
    be combined with axioms that split larger numbers to perform long multiplication.
    In \texttt{fractions}, operations include factoring numbers
    into a prime multiplied by a divisor, canceling common factors
    and combining fractions with the same denominator.
    Finally, in \texttt{sorting}, the agent needs to sort
    delimited substrings by length by either reversing the entire list
    or performing adjacent swaps. 
    }
    \label{fig:domain-examples}
\end{figure}

To overcome these challenges, we propose a novel learning algorithm, Contrastive Policy Learning (ConPoLe), which succeeds in symbolic environments.
Our key insight is to directly learn a policy by attempting to capture the mutual information between current and future states that occur
in successful trajectories. 
ConPoLe uses iterative deepening and
beam search to find successful and failed trajectories
during training.
It then uses these positive and negative examples to optimize the InfoNCE loss \cite{oord2018representation}, which lower bounds the mutual information between
the current state and successful successors.
This provides a new connection between policy learning and unsupervised contrastive learning.
Our main contributions in this paper are:

\begin{itemize}
    \item We introduce 5 environments for symbolic reasoning (Fig.~\ref{fig:domain-examples})
    drawn from skills listed in the Mathematics Common Core Curriculum (Section~\ref{sec:setup}).
    We find that
    existing Reinforcement Learning algorithms fail to
    solve these domains. 

    \item We formulate policy learning in deterministic environments
    as contrastive learning,
    allowing us to sidestep value estimation (Section~\ref{sec:solution}).
    The algorithm we introduce, ConPoLe, succeeds in all five Common Core environments, as well as in solving the Rubik's Cube (Section~\ref{sec:eval}).

    \item We provide quantitative and qualitative evidence
    that the problem representations learned by
    ConPoLe reflect the equation-solving curriculum
    from the Khan Academy platform.
    This result suggests a number of applications of representation learning in education.
\end{itemize}

\section{Related Work}

Automated mathematical reasoning spans several research communities.
Theorem-proving languages, such as Coq \cite{barras1997coq} or Lean \cite{de2015lean},
enable the formalization of mathematical statements and can verify proofs,
but they are limited in their ability to \emph{discover} solutions automatically.
A rich line of recent work has focused on learning to produce formal
proofs using supervised \cite{polu2020generative, yang2019learning, alemi2016deepmath} and reinforcement learning \cite{bansal2019holist, kaliszyk2018reinforcement, piotrowski2018atpboost}.
Similar to GPT-f \cite{polu2020generative}, our model only assumes
unstructured strings as input; however, we do not use a dataset of human solutions.
To the best of our knowledge,
our work is the first to consider learning formal reasoning directly
from text input (like GPT-f) by purely interacting with an environment
(like rlCop \cite{kaliszyk2018reinforcement} and ASTatic \cite{yang2019learning}).

Existing RL algorithms for theorem proving make use of partial rewards to guide
search. For example, rlCop \cite{kaliszyk2018reinforcement} learns to prove in the Mizar Verifier using Monte Carlo Tree Search.
In Mizar, applying a theorem decomposes the proof into sub-goals. Random
policy rollouts often close a fraction of the sub-goals, providing signal to MCTS.
In our environments, however, there is only one solution state for each problem,
causing MCTS to degenerate to breadth-first search.
This challenge is similar in other large search problems,
such as the Rubik's Cube. DeepCubeA \cite{agostinelli2019solving}
handles the sparsity problem in the Rubik's Cube by generating
examples in reverse, starting from the solution state.
This uses the fact that moves in the Rubik's Cube have
inverses, and that the solution state is always known a priori.
In contrast, the algorithm we propose in Section~\ref{sec:solution},
which has neither of these assumptions, is still able to learn an effective Rubik's Cube solver; we compare it to DeepCubeA in Section~\ref{sec:eval}.

% Classical tutors were reported to typically use hundreds of programmer-written rules to find solutions in domains such as algebra or SQL queries\cite{r1991development}, making them notoriously expensive to develop\cite{aleven2009new}. The challenge remains in recent work: in practice, developers of modern automated tutors either resort to simple search-based solvers, which typically can only handle short problems, or manually write complex solvers equipped with domain-specific planning heuristics. An example of the former case is the Algebra Notepad \cite{o2019automatic}, a recently proposed automated tutor that is limited to solution of up to 4 steps due to its backtracking solver. On the other hand, Google's Socratic educational app utilises a hand-written solver for equations and fractions, MathSteps, which spans 7,000 lines of code, mostly implementing heuristics about when to apply each available action. This is necessary to find longer solutions because the search space of action sequences quickly makes search intractable\footnote{In our implementation of a step-by-step equation solver, when considering all actions that are applicable in each subterm (e.g. commutativity, associativity, evaluation, applying simultaneous operations to both sides, etc), roughly 27 choices will be available on average in a given equation, with harder equations requiring up to 30 steps.}

Intelligent
Tutoring Systems \cite{ma2014intelligent, ritter2007cognitive, mccalla1992search} provide a key application for step-by-step solvers. Tutors for symbolic domains, such as algebra \cite{ritter2007cognitive, o2019automatic} or database querying \cite{anderson1985intelligent}, can use a solver
to help students even in the absence of expert human hints
(unlike \emph{example-tracing tutors}, which simply follow extensive problem-specific
annotations \cite{weitekamp2020interaction}).
Currently, solvers used in automated tutors are either
search-based \cite{o2019automatic}, but limited to short solutions, or hand-engineered \cite{mathsteps}.
Our method can learn high-level solution strategies
by composing simple axioms, generalizing to long solutions without
the need for expert heuristics. Moreover, our analysis in
Section~\ref{sec:solution} suggests that learned problem embeddings can be
applied in mathematics education -- another advantage of a neural solver.

Finally, our approach to Reinforcement Learning in symbolic domains builds on
unsupervised contrastive learning, specifically InfoNCE \cite{oord2018representation}. Contrastive learning
has been used to learn representations that are then used by existing Deep RL algorithms \cite{oord2018representation, laskin2020curl, fu2020towards}.
We instead make a novel connection,
casting policy learning as contrasting examples of successful and failed trajectories.

% [NDG: i moved this from beggining, but some of it probably should be incorportaed?

%In contrast, \emph{Artificial} Intelligence systems, and in particular
%deep neural networks, have been recently much more successful
%in tasks associated with \emph{perception}
%(e.g. image classification and speech recognition) than reasoning.
%Even tasks that are reasoning-intensive for humans, such as (visual) question
%answering, tend to be solved by neural networks with a combination of accurate
%perception and dataset correlations %\cite{agrawal2016analyzing,goyal2017making,agrawal2018don}.
%While neuro-symbolic models which explicitly incorporate reasoning components,
%have recently received notable attention %\cite{pmlr-v119-amizadeh20a,mao2018neuro,DBLP:journals/corr/ParisottoMSLZK16,dang2020plans,valkov2018houdini}, automatically learning general
%reasoning skills remains a largely open problem.
%]

\section{Setup}
\label{sec:setup}

\begin{table}
    \centering
    \caption{Common Core environments for symbolic reasoning.}
    \footnotesize
    \label{tab:environments}
    \begin{tabular}{@{}l l l l@{}}
        \toprule
        Environment & Reference & Average & BFS \\
        &  & Branching Factor & Success Rate \\
        
        \midrule
        sorting & CCSS.Math.Content.MD.A & 5.84 & 17\% \\
         & Sort objects by length. & & \\
         
        ternary-addition & CCSS.Math.Content.1.NBT.C & 10.55 & 9\% \\
         & Perform step-by-step arithmetic with carry. & & \\
         
        multiplication & CCS.Math.Content.1.NBT.C & 8.62 & 26\% \\
        & Multiply multi-digit whole numbers & & \\

        fractions & CCSS.Math.Content.NF.B & 6.58 & 18\% \\
         & Manipulate expressions with fractions. & & \\

        equations & CCSS.Math.Content.8.EE.C & 27.12 & 5.5\% \\
         & Solve linear equations in one variable. & & \\
        \bottomrule
        \addlinespace{}
    \end{tabular}
\end{table}

Motivated by mathematics education, we are interested in formal domains
where problems are solved in a step-by-step fashion, such as equations
or fraction simplification. While various algorithms could compute the final
solution to such problems, finding a \emph{step-by-step} solution by
composing low-level axioms is a planning problem. Formally,
we define a \emph{domain} $\mathcal{D} = (S, A, T, R)$ as a deterministic
Markov Decision Process (MDP), where $S$ is a set of states, $A(s)$ is the set of
actions available at state $s$, $T(s_t, a)$ is the transition function,
deterministically mapping state $s_t \in S$ and action $a \in A(s_t)$ to the
next state $s_{t+1}$, and $R(s)$ is the binary reward of a state:
$R(s) = 1$ if $s$ is a ``solved'' state, otherwise $R(s) = 0$.
States with positive reward are terminal.
Initial states can be sampled from a probability distribution $P_I(s)$,
which we assume to have states of varying distance
to the solution state. This implicit curriculum allows agents
to find a few solutions from the beginning of training -- an important starting signal,
since we assume no partial rewards.

States in symbolic domains typically have a well-defined underlying structure.
For example, mathematical expressions, such as equations, have a recursive
tree structure, as do SQL queries and programming languages.
A different, yet similarly well-defined structure dictates valid states
in the Rubik's cube.
However, in order to study the general problem of symbolic learning, we want to assume no structure for states beyond the MDP specification. 
Therefore, we assume all states
and actions are strings, i.e. $S, A(s) \subseteq \Sigma^*$
for some alphabet $\Sigma$. Naturally, our goal is to learn a policy
$\pi(a|s)$ that maximizes the probability that following $\pi$ when starting at 
a randomly drawn state $s_0 \sim P_I(\cdot{})$ leads to a solved state $s_k$.

\subsection{Environments}

We introduce four environments that exercise core abilities listed in the Mathematics
curriculum of the Common Core State Standards. The Common Core is an initiative
that aims to build a standard high-quality mathematics curriculum
for schools in the United States, where ``forty-one states, the District of Columbia, four territories, and the Department of Defense Education Activity (DoDEA) have voluntarily adopted and are moving forward with the Common Core''\footnote{\url{http://www.corestandards.org/about-the-standards/}}. We draw inspiration from four key contents
in the curriculum: ``Expressions and Equations'', ``Numbers and Operations -- Fractions``, ``Measurements and Data'', and ``Operations and Algebraic Thinking''.

Table~\ref{tab:environments} lists these environments.
To put their respective search problems into perspective, we report two statistics:
the average branching factor of $1M$ sampled states,
and the success rate of a simple breadth-first search (BFS) in solving 1000
sampled problems, when limited to visiting $10^7$ state-action edges.
All the environments require strategy and exploiting domain structure to be
solved: na\"ive search succeeds only on the simplest problems,
with success rates ranging from $26\%$ in \texttt{multiplication} to
$5.5\%$ in \texttt{equations}. We describe all axioms
available in each environment in detail in the Appendix. 
Figure~\ref{fig:domain-examples} gives an example of a problem and
step-by-step solution in each of the environments.
Agents operate in a fairly low-level axiomatization of each
domain: simple actions need to be carefully applied in sequence to perform
desirable high-level operations. For example, to eliminate an additive term
in one side of an equation, one needs to subtract that term from both sides,
rearrange terms using commutativity and associativity to obtain a sub-expression
with the term minus itself, and finally apply a cancelation axiom that produces zero.
This formulation allows agents to compose actions to form general strategies,
but at the same time makes planning challenging.

\section{Contrastive Policy Learning}
\label{sec:solution}

A common paradigm in Reinforcement Learning uses rollouts to learn value estimates: a starting state is sampled, the agent executes its
current policy until the trajectory horizon, and state value is updated using
observed rewards.
However, when rewards are
sparse, as in the symbolic environments considered here, this paradigm suffers from a serious
\emph{attribution problem}. In failed trajectories, which are typically the
vast majority encountered by an untrained agent, the signal of a null reward
is weak:
failure could have happened due to any of the steps taken.
Moreover, the agent has no direct feedback on what
\emph{would have happened} had it taken a different action in a certain state, until it visits another similar state.
In turn, long intervals between similar observations
can hinder leaning.
These issues are ameliorated by Monte Carlo Tree Search, and similar algorithms, which explore a search tree and may abandon
a sub-tree based on expected outcomes.
However, MCTS relies on Monte Carlo estimates of the value of candidate states.
In games, such as Go or Chess, random rollouts find a terminal state per problem, providing reward signal. (Indeed, self-play in such games always terminates and wins about half of games.) But in domains with a single terminal state,
like math problems or the Rubik's cube, such estimates are
unhelpful \cite{agostinelli2019solving},
returning zero reward with probability that approaches $1$ exponentially in the
distance to the solution.

To obtain signal about each step of a solution, we would like to
consider multiple alternative paths, as MCTS does, without relying on
random rollouts to yield useful value estimates. To that end, we employ \emph{beam search}
using the current policy $\pi$, maintaining a beam of the $k$ states
most likely to lead to a solution. When a solution is found,
the beam at each step contains multiple alternatives to the action that
eventually led to a solution. We take those alternatives as
\emph{negative examples}, and train a policy that maximizes the probability of picking the successful decision over alternatives.

Let $f_\theta(p, s_t)$ be a parametric function that assigns a non-negative score
between a proposed next state $p$ and the current state $s_t$.
Although any non-negative function suffices, here we use a simple 
log-bilinear model that combines the embeddings of the current
and proposed states, both obtained with the state embedding function
$\phi_\theta : S \rightarrow \mathbb{R}^n$:

$$f_\theta(p, s_t) = \exp\left(\phi_\theta(p)^\top \cdot{} W_\theta \cdot{} \phi_\theta(s_t) \right)$$

Once we find a solution using beam search, for each intermediate step $t$ we
have a state $s_{t}$ and a set $X = \{ p_1, \cdots, p_N \}$ of next state proposals, obtained from actions that were available at states in the beam at step $t$. One of these, which we call $s_{t+1}$, is the proposal
in the path that led to the found solution. 
We then minimize:

$$\mathcal{L}_t(\theta) = \mathbb{E}_{s_t} \left[ - \log \frac{f_\theta(s_{t+1}, s_t)}{\sum_{p_i \in X}f_\theta(p_i, s_t)}\right]$$

Algorithm~\ref{alg:conpole} describes our method, which we call Contrastive
Policy Learning (ConPoLe). In successful trajectories, ConPoLe
adds positive and negative examples of actions to a
\emph{contrastive experience replay buffer}.
After each problem, it samples
a batch from the buffer and optimizes $\mathcal{L}_t$.
Since we can only find positives when a solution is found, ConPoLe
essentially ignores unsolved problems.
Moreover, to improve data efficiency, ConPoLe iteratively deepens
its beam search: with an uninitialized policy, it cannot expect to find
long solutions, so exploring deep search trees is unhelpful.
In our implementation, we simply increase the maximum search depth
by $1$ every $K$ problems solved, up to a fixed maximum depth.

%Thus, we can show that the optimal $f$will be proportional to a likelihood ratio: $f(s_{t+1}, x_{t}) \propto \frac{P(s_{t+1}|s_t)}{P(s_{t+1})} \enspace .$
Our loss $\mathcal{L}_t$ is equivalent to the InfoNCE loss, which is shown in \cite{oord2018representation} to bound the mutual information: $I(s_{t+1}, s_t) \geq \log |X| - \mathcal{L}_t \enspace .$
Note that in our domains each state is the next step for the correct solution to \textit{some} problem, thus the negative set found by beam search approximates negative samples of next states from other solutions, though with a bias toward closer `hard negatives'\cite{wu2021conditional}.
Thus our approach can be interpreted as learning a representation that captures the mutual information between current states and their  successors along successful solutions.
The MI bound becomes tighter with more negatives:
we explore this property experimentally in Section~\ref{sec:eval}.
We refer to \cite{oord2018representation} for a detailed derivation of this bound and its properties.

As our policy we directly use the similarity function $f(p_{i}, s_t)$ normalized over possible next states.
The objective $\mathcal{L}_t$
is also the categorical cross entropy between the model-estimated next state
distribution $\pi_\theta(p|s_t) \frac{f(p, s_t)}{\sum_{p_j \in A(s)}f(p_j, x_t)}$ and the distribution of successful proposals. It is thus minimized when the predictions match $P(s_{t+1} = p_i|s_t)$, the optimal policy. 

% by having
% negatives to contrast to the successful action, we can directly learn
% what action to take to maximize the probability of finding a solution --
% exactly the role of a policy, which is usually obtained by more indirect
% methods.

The ConPoLe approach avoids value estimation by focusing directly on statistical properties of solutions.
The \texttt{sorting} environment illustrates the intuition
that this is a simpler objective. In this domain, the agent has to sort
a list of substrings by their length, using adjacent swaps or
reversing the whole list. The value of a state would be proportional to the number of inversions (i.e. pairs of indices that are in the incorrect order). This prediction problem is hard for learning: $O(n^2)$ pairs of indices need to be
considered\footnote{A $\Theta(n\log n)$ solution
counts inversions by exploiting the merge sort recursion  \cite{cormen2009introduction}.}.
ConPoLe, however, only needs to tell if a proposed state has more or less inversions than the current
state. In the case of adjacent swaps, this amounts to detecting whether the swapped elements were previously in the wrong order.
By having negative examples to contrast with successful trajectories,
ConPoLe can learn a policy by completely sidestepping value estimation.

% Beyond intuition, its connection to unsupervised
% Contrastive Learning yields a principled interpretation of ConPoLe's
% objective: it seeks to preserve mutual information between
% the current state and successors most likely to
% be observed in a successful trajectory.

\RestyleAlgo{ruled}
% \LinesNumbered

\begin{algorithm}

\DontPrintSemicolon
\KwIn{Environment $E$}
\KwOut{Learned policy parameters $\pi_\theta$}
$\theta \gets \texttt{init\_parameters}()$\;
$\mathcal{D} \gets \varnothing$\;
$n\_solved \gets 0$\;
\For{$episode \gets 1$ \textbf{to} $N$} {
  $p \gets E.\texttt{sample\_problem()}$\;
  $(solution, visited\_states) \gets \texttt{beam\_search}(E, \pi_\theta, p,
  beam\_size, max\_depth)$\;
  \If{$solution \neq \texttt{null}$} {
    $n\_solved \gets n\_solved + 1$\;
    $neg\_states \gets visited\_states \setminus solution$\;
    \For{$i \gets 1$ \textbf{to} $\texttt{length}(solution) - 1$} {
        $pos \gets (solution[i], solution[i+1])$\;
        $neg \gets \{(solution[i], c) : c \in neg\_states \texttt{ from step i of beam\_search} \}$\;
        $\mathcal{D}.\texttt{add}\left(\langle pos, neg\rangle\right)$
    }
    $B \gets \mathcal{D}.\texttt{sample\_batch}$()\;
    $\theta \gets \theta - \alpha \nabla \texttt{InfoNCE}(\theta, B)$
  }
}

\caption{Contrastive Policy Learning (ConPoLe)}
\label{alg:conpole}
\end{algorithm}

\section{Experiments}

We now evaluate our method guided by the
following research questions:
How does ConPoLe perform when solving symbolic reasoning problems from educational domains?
How do negative examples affect ConPoLe's performance?
Are ConPoLe's problem embeddings useful for downstream educational applications?
Can ConPoLe be applied to other large-scale symbolic problems?

\subsection{Setup}

We compare ConPoLe against four RL baselines and the
Google MathSteps\footnote{\url{https://github.com/google/mathsteps}} 
library, which contains manually implemented step-by-step
solvers for the Fractions and Equations domains (as opposed to simply
giving the final answer, as several other libraries do).
The first learning-based baseline is the
\emph{Deep Reinforcement Relevance Network} \cite[DRRN]{he2016deep},
an adaptation of Deep Q-Learning for environments with textual
state and dynamic action spaces.
We additionally test Autodidatic Iteration \cite[ADI]{mcaleer2018solving}
and Deep Approximate Value Iteration \cite[DAVI]{agostinelli2019solving} 
-- both methods have been recently used to solve the Rubik's Cube,
a discrete puzzle that is similar to the Common Core domains in that
it only has a single solution state.
Finally, we use a simple Behavioral Cloning (BC) baseline, as done in
\cite{chen2021decision}: it executes a random policy until its
budget is exhausted; then, it trains a classifier only on the successful
trajectories that picks the successful action.

As a simpler alternative to ConPoLe, we use a baseline we call
\emph{Contrastive Value Iteration} (CVI). This algorithm
is identical to ConPoLe except for its loss:
we train it to predict the final reward obtained by each explored state.
In other words, after finding a solution,
CVI will add examples to its replay buffer of the form
$(s, r) \in \mathcal{S} \times \mathbb{R}$,
where $s$ is a visited state and $r$ is $1$ if that state
occurred in the path to the solution, or $0$ otherwise.
This can be seen as estimating $P(s')$: the probability
that $s'$ would be observed in a successful trajectory,
without conditioning on the current state. Since the reward
is binary, this corresponds to value estimation.

All models use two-layer character-level LSTM \cite{hochreiter1997long} network to encode inputs (DRRN uses two such networks).
Each agent was trained for $10^7$ steps in each environment; runs took from 24 to 36 hours on a single NVIDIA Titan Xp GPU.
Problems in the \texttt{equations} domain come from a set
of 290 syntactic equation templates (with placeholders for constants, which
we sample between -10 and 10.) extracted from the Cognitive Tutor Algebra \cite{ritter2007cognitive} dataset. Other environments
use generators we describe in the Appendix.
Code for agents and CommonCore environments is available at \url{https://github.com/gpoesia/socratic-tutor}.
Our Common Core environments are implemented in Rust,
and a simple high-throughput API is available for Python.

\subsection{Solving CommonCore domains}

\label{sec:eval}

\begin{table}
%\small
    \centering

\begin{tabular}{l c c c c c}
\toprule
Agent & Sorting & Addition & Multiplication & Fractions & Equations\\
\midrule
BC & 92.0\% & 64.5\% & 10.0\% & 52.5\% & 4.5\%\\
DRRN & 29.5\% & 40.0\% & 10.5\% & 20.0\% & 2.5\%\\
ADI & 91.0\% & 63.5\% & 9.5\% & 46.5\% & 5.5\%\\
DAVI & 99.5\% & 54.0\% & 18.5\% & 57.5\% & 8.0\%\\
\emph{MathSteps$^{(N)}$} & - & - & - & 100\% & 76.0\%\\
\midrule
CVI & 77.0\% & 72.0\% & 100.0\% & 84.5\% & 89.0\%\\
ConPoLe-local & 100.0\% & 98.5\% & 99.5\% & 86.0\% & 76.5\%\\
ConPoLe & 100.0\% & 100.0\% & 99.5\% & 96.0\% & 92.5\%\\
\bottomrule
\end{tabular}

    \caption{Success rate of all agents in the CommonCore environments.
    Agents were ran with 3 random seeds for $10^7$ environment steps,
    and tested every $100k$ steps on a held-out set of $200$ problems.
    We report the best observed success rate of each agent's
    greedy policy (i.e. no search is done at test time).
    $(N)$ emphasizes that the MathSteps library is not learning-based: we simply ran it on test problems.
    }
    \label{fig:main-result}
\end{table}

We start by comparing agents when learning
in the CommonCore environments.
In this experiment, agents train using sequentially sampled random problems.
We define one \emph{environment step} as a query in which the agent specifies a state,
and the environment either returns that the problem is solved or lists available actions
and corresponding next states.
Since trajectories can be potentially infinite, we limit agents to search for a maximum depth of $30$,
which was enough for us to manually solve a sample set of random training problems from all environments.

Table~\ref{fig:main-result} shows the best performance of each agent
on a held-out test set of $200$ problems.
Success rate is measured using each agent's greedy policy: we do not perform
any search at test-time.
DRRN fails to effectively solve any of the environments.
ADI, DAVI and BC can virtually solve the sorting domain,
but this performance does not translate to harder domains multiplication and
equations. ConPoLe shows strong performance on all domains, with CVI being
comparable in equations and multiplication, but falling behind in the other two domains.

DRRN quickly converges to predicting near $0$ for most state-action pairs,
failing to learn from sparse rewards. We note that the environments where DRRN has shown success, such as text-based games, have shorter states and actions as well as intrinsic correlations between states (derived from natural language). These features may help to smooth the value estimation problem but are not available in the Common Core domains.

We note that ADI and DAVI have been previously successful in puzzles with an important feature:
the state sampler produces problems that are exactly
$k$ steps from the solution for all $k$ up to the maximum necessary.
For instance, in the Rubik's Cube, scrambled cubes can be generated by
starting from the solution state and performing $k$ moves in reverse.
This feature is present in the \texttt{sorting} domain (in which both ADI and DAVI perform well),
but not in general.
For example, even the simplest equation in the Cognitive Tutor dataset
still requires 3 actions to be solved; some require up to 30, and there are many gaps in this progression.
These gaps cause ADI and DAVI to find states at test time that are out of their training distribution.
As suggested in \cite{mcaleer2018solving}, we experimented with replacing the 1-step lookahead of these algorithms by a bounded BFS. However, we found that this na\"ive exploration strategy
is sample inefficient and did not significantly change their performance.

% where states and actions were much shorter
% (e.g. two-word actions in their parser-based game).
% Moreover, in natural language descriptions, intrinsic
% correlations between words and state-action values might help
% with learning. This is not the case in our symbolic
% environments. For instance, in the equations domain, the
% same action of dividing both sides by $2$ will
% be available in many similar-looking states,
% yet it only moves the agent closer to the solution in few situations.
% For example, in the equations $2 + x = 4$, $4x = 2$, $2x + 2 = 4$, $2x = 4$,
% dividing by $2$ is only useful in the last case.

On the other hand, algorithms that learn from contrasting
examples (CVI and ConPoLe) perform well in all environments.
Using only examples derived from solved problems
gives much higher signal to each data point: we only consider
a ``negative'' when we have a corresponding positive example
which actually led to a solution.
Moreover, we observe gains in using the InfoNCE loss for
training with contrasting examples: ConPoLe
performs consistently better than CVI, and
can very quickly learn to solve almost all sorting and addition problems.
In these domains, deciding whether an action moves towards
the goal is an easy problem, but precisely estimating values is challenging.
In Sorting, for example, CVI's learned policy becomes unreliable for lists with more than 7
elements. We observe a similar limitation in Addition,
where one sub-problem is sorting digits by the power of 10 they multiply.

These observations provide evidence to answer our first
research question: using contrastive examples is beneficial to
learning policies in symbolic domains, and explicitly optimizing a contrastive
loss improves results further.

\paragraph{The impact of negative examples}
Contrastive learning algorithms have been observed
to perform better with more negative samples: the variance of the InfoNCE estimator decreases with more negative examples.
However the choice of negative examples can also impact the performance \cite{wu2021conditional}.
We thus experimented with
a simple variant of ConPoLe that produces fewer, but more local, negative examples.
Instead of using \emph{all} candidates visited by beam
search,
we instead only use those that were actual candidates
for successors of states in the solution path.
We call this variant \emph{ConPoLe-local}, since it only
uses ``local'' negatives. The performance of ConPoLe-local is shown in Table~\ref{fig:main-result}. ConPoLe-local behaves
like ConPoLe in Sorting, Addition and Multiplication.
In the last two domains, however,
there is a significant performance gap between
ConPoLe and ConPoLe-local.
Interestingly, this does not seem to be the case for CVI.
We executed the same experiment with CVI, and observed
no reliable difference in performance or learning behavior.
This finding supports the importance of the connection between reinforcement learning and contrastive learning: using all available negatives, in the way suggested by InfoNCE, yields the best results. 

\subsection{Comparing learned representations to human curricula}

\begin{wrapfigure}{r}{0.5\textwidth}
%\begin{figure}
\centering
%\begin{floatrow}
%\ffigbox{%
    \includegraphics[width=0.48\textwidth]{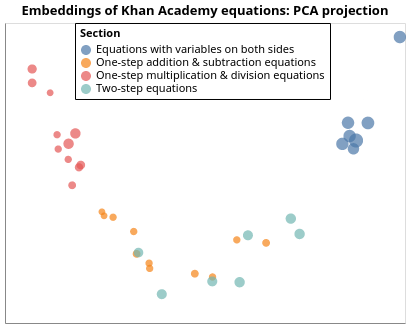}
%}{%
  \caption{PCA projection of ConPoLe's learned
           representations for equations from Khan Academy. Point sizes indicate ConPoLe's solution length
           (ranging from 7 to 29 in these exercises).
           We observe clusters that closely match the different sections
           of the Khan Academy curriculum.}%
   \label{fig:pca}
\end{wrapfigure}

A learned neural solver yields distributed representations for problems.
One natural question is whether these representations capture properties
of the problems that resemble how humans structure the same domains.

To investigate this question, we collected a dataset of equations from the Khan Academy\footnote{https://www.khanacademy.org/} educational portal.
We used the equations listed as examples and test problems
in the sections of the algebra curriculum dedicated to linear
equations. There are four sections:
``One-step addition and subtraction equations'',
``One-step multiplication and division equations'',
``Two-step equations'' and
``Equations with variables on both sides''.
The first three categories had 11 equations, and the last had 9.

We evaluated a $1$-nearest neighbor classifier
that predicts the equation's category from the closest labeled example, based on four distance metrics:
string edit distance (a purely syntactical baseline),
and cosine similarity using the representations learned
by agents. Chance performance in this task is $26\%$.

Table \ref{tab:khan-academy} shows ConPoLe's representations
yield accurate category predictions (90.5\%),
while other representations
are less predictive than the Edit Distance baseline.
We further observe that ConPoLe's latent space is highly structured
(Figure~\ref{fig:pca}): sections from Khan Academy form visible clusters.
This happens despite of ConPoLe being trained without an explicit curriculum or examples of human solutions.
%This observation suggests a number of applications of
%learned representations for future work,
%such as their use in adaptive problem suggestions \cite{srivastava2021question, mu2021automatic} and
%in amortized item-response theory \cite{wu2020variational}.

% having never been given an
% explicit curriculum, and even if its learned solution strategies
% often differ from how humans solve equations.
% For example, we observed that ConPoLe has a tendency
% to leave trivial operations to the very end: it will
% often carry around a ``+ 0'' for many steps, after
% canceling two terms. Humans, on the other hand, tend to simplify early.
% Nevertheless, we were able to use ConPoLe's representations
% to accurately match exercises in the Khan Academy curriculum.

\begin{table}
    \centering
    \caption{Accuracy of predicting equation categories from learned representations on Khan Academy.}
      \begin{tabular}{c | c c c c c c c} \toprule
      \textbf{Representation} & BC & DRRN & ADI & DAVI & CVI & Edit Distance & ConPoLe \\
      \textbf{Accuracy} & 0.619 & 0.642 & 0.619 & 0.619 & 0.642 & 0.714 & \textbf{0.905} \\
      %\midrule
      %DRRN & 0.642 \\
      %ADI & 0.621 \\
      %DAVI & 0.621 \\
      %CVI & 0.642 \\
      %Edit Distance & 0.714 \\
      %ConPoLe & \textbf{0.905} \\
      \bottomrule
      \end{tabular}
      \label{tab:khan-academy}
\end{table}

\subsection{Solving the Rubik's Cube}

Finally, we apply ConPoLe to a challenging search problem: the Rubik's Cube \cite{agostinelli2019solving}.
This traditional puzzle consists of a 3x3x3 cube, with 12 possible
moves that rotate one of the 6 faces either clockwise or
counterclockwise. 
Initially, all 9 squares in each face have
the same color. The cube can be scrambled by applying face rotations,
and the goal is to bring it back to its starting state.
There are $4.3 \times 10^{19}$ valid configurations of the cube,
and a single solution state.
We compare ConPoLe to DeepCubeA \cite{agostinelli2019solving},
a state-of-the-art solver that learns with Deep Approximate Value Iteration.

For this task, we simply represent the
cube as a string of 54 digits from $0$ to $5$, representing 6 colors,
with a separators between faces.
We run the exact same architecture we used in the Common Core domains.
We train ConPoLe for $10^7$ steps on a single GPU,
with a training beam size of $1000$, on cubes scrambled with up to
20 random moves. 
(DeepCubeA observed 10 billion cubes during training
time, compared to 10 million environment steps taken by ConPoLe during training.)
In test time, following DeepCubeA,
we employ Batch Weighted A* Search (BWAS),
using our model's predicted log-probabilities as weight estimates.

We find that ConPoLe is able to learn an effective solver.
We tested on 100 instances scrambled with 1000 random moves, as used in DeepCubeA's evaluation. ConPoLe succeeds in all cubes (as does DeepCubeA).
In finding solutions, BWAS paired with ConPoLe visits an average of 3
million nodes, compared to 8.3 million with DeepCubeA.
DeepCubeA solutions however are shorter (21.6 moves on average, compared to ConPoLe's 39.4).\footnote{BWAS can be tuned to trade off shorter solutions for exploring more nodes;
we only tested the default parameters in the DeepCubeA implementation.
Given DeepCubeA was trained on 1000x more cubes, we would expect
its solutions to still be shorter if we matched both algorithms in number
of visited nodes.
}
Overall, this result validates the generality and promise of ConPoLe for solving challenging symbolic domains.

\section{Conclusion}
\label{sec:conclusion}

We introduced four environments inspired by mathematics education,
in which Reinforcement Learning is challenging.
Our algorithm, based on optimizing a contrastive loss, demonstrated
significant performance improvements over baselines.
While we used educational domains as a test bed,
our method can in principle be applied to any discrete planning domain with binary rewards.
One requirement is that an untrained agent must find enough
solutions to assemble initial contrastive examples.
Procedural mathematical domains are a natural instance of these.
However, our educational environments assume a linear structure,
where applying an axiom directly leads to the next state.
This assumption breaks in more expressive formulations of formal mathematics (such as first-order logic or dependent type theories), where proofs have a tree structure.
Other domains, such as programs, might pose additional
challenges because an unbounded number of actions can be available at a state.
Adapting ConPoLe for these settings is non-trivial, and poses important challenges for future work.

Our learned solvers can simplify the process
of building Intelligent Tutoring Systems. These systems can free educators to focus on more conceptual problems.
On the down side, solvers can
provide unfair resources for homework and unequal access could exacerbate inequality. 
Beyond tutoring, we observed that the representations
learned by our agents capture semantic properties about problems.
This opens up an avenue for additional research on
deep representations for educational applications.

\section*{Acknowledgements}

We thank the anonymous NeurIPS reviewers for the valuable discussion, which significantly improved our work.
This work was supported by a NSF Expeditions Grant, Award Number (FAIN) 1918771.

\bibliography{references.bib}

\appendix

\section{Common Core environments}
\label{sec:ap-commoncore}

In Section~\ref{sec:setup}, we briefly described four Common Core-inspired
environments: \texttt{equations}, \texttt{fractions},
\texttt{ternary-addition} and \texttt{sorting}.
We now provide a detailed description of the states, actions and
problem generators for each of these environment.

\subsection{\texttt{equations}}

The \texttt{equations} environment exercises the ability to
coordinate primitive algebraic manipulations in order to solve an equation.
Each problem is a linear equation on a single variable $x$,
and actions are valid manipulations of the equation, following simple axiomatic rules.
A valid state in this domain is an \emph{equality},
which is comprised of two \emph{expressions}: one on the left
and one on the right. In turn, an expression can be one of the following:

\begin{description}
    \item[Constant: ]An integer $n$, or a rational $\frac{a}{b}$ with $b \neq 0$,
    \item[Binary operation: ]A (recursively defined) left-hand side expression $e_l$, an operator $op \in \left\{+, -, \times, / \right\}$, and a right-hand side expression $e_r$,
    \item[Unary operation: ]The operator $-$ followed by an expression $e_r$,
    \item[Unknown: ] The unknown $x$.
\end{description}

A state is solved only when it is in the form $x = n$, where $n$ is a constant.
When representing states as strings, we use the standard mathematical notation,
with the detail that we parenthesize all binary operations so that operator precedence is made explicit.

To generate problems in this domain, we leverage the Cognitive Tutor Algebra
dataset \cite{ritter2007cognitive}. This dataset contains logs of student
interactions with an automated algebra tutor. We collected all equations from
the logs, and replaced their numerical constants by placeholders.
This gave us 290 syntactic equation templates, such as
$$(\msquare{}x + \lozenge) = x$$ and
$$(\msquare - (-\lozenge)) = (((\bigstar / x) + (-\maltese)) - (-\blacklozenge)) \enspace .$$

To generate a problem, we first sample one of the templates, and then
replace each constant independently by an integer between -10 and 10 inclusive,
uniformly.

Table~\ref{tab:eq-actions} lists all axioms in the domain, with examples
of applying each.

\begin{table}
    \centering
    \caption{Axioms of the \texttt{equations} domain.}
    \label{tab:eq-actions}
    \begin{tabular}{p{1.7cm} p{4.8cm} p{6cm}}
        \toprule
        Mnemonic & Description & Example \\
        \midrule
        refl & Reflexivity: if $a = b$, then $b = a$. & $1 + 2 = x \rightarrow x = 1 + 2$ \\
        comm & Commutativity: $+$ and $\times$ commute. & $(2x) / 2 = 4 \rightarrow (x \times 2) / 2 = 4$ \\
        assoc & Associativity: $+$ (resp. $\times$) associates over $+$ and $-$ (resp. $\times$ and $/$). & $((x + 1) - 1) = 9 \rightarrow (x + (1 - 1)) = 9$ \\
        dist & Distributivity: $\times$ and $/$ distribute over $+$ and $-$. & $2 \times(x + 1) = 5 \rightarrow (2x + (2 \times 1)) = 5$ \\
        sub\_comm & Consecutive subtractions can have their order swapped. & $((2x - 1) - x) = 1 \rightarrow ((2x - x) - 1) = 1$ \\
        eval & Operations with constants can be replaced by their result. & $x = (9 / 3) \rightarrow x = 3$ \\
        add0 & Adding $0$ is an identity operation. & $(x + 0) = 9 \rightarrow x = 9$ \\
        sub0 & Subtracting $0$ is an identity operation. & $(x - 0) = 9 \rightarrow x = 9$ \\
        mul1 & Multiplication by $1$ is an identity operation. & $1x = 9 \rightarrow x = 9$ \\
        div1 & Division by $1$ is an identity operation. & $(x / 1) = 9 \rightarrow x = 9$ \\
        div\_self & Dividing a non-zero term by itself results in $1$ & $x = (5x / 5x) \rightarrow x = 1$ \\
        sub\_self & Any term minus itself is $0$. & $x = ((x + 1) - (x + 1)) \rightarrow x = 0$ \\
        subsub & Subtracting $-e$ is equivalent to adding $e$. & $(x - (-9)) = 10 \rightarrow (x + 9) = 10$ \\
        mul0 & Multiplying by $0$ results in $0$. & $x = (1 + 0\times2x) \rightarrow x = (1 + 0)$ \\
        zero\_div & $0$ divided by a non-zero term results in $0$. & $x = (0 / (x+1)) \rightarrow x = 0$ \\
        add & Any subterm can be added to both sides of the equation. & $(x - 1) = 0 \rightarrow ((x - 1) + 1) = (0 + 1) $ \\
        sub & Any subterm can be subtracted from both sides of the equation. & $(x + 1) = 0 \rightarrow ((x + 1) - 1) = (0 - 1) $ \\
        mul & Any subterm can be multiplied to both sides of the equation. & $(x/2) = 6 \rightarrow ((x/2)\times2) = (6\times2)$ \\
        div & Any subterm can be used to divide both sides of the equation. & $2x = 6 \rightarrow ((2x)/2) = (6/2)$ \\
        \bottomrule
    \end{tabular}
\end{table}

The following are two examples of step-by-step solutions generated by ConPoLe
for sampled problems, with the axioms used to derive each step.
Numbers in square brackets represent fractions, not divisions (e.g. $\texttt{[4/5]}$ means $\frac{4}{5}$).

\begin{footnotesize}
\begin{verbatim}
(-7) = (3 - ((-7) / x)) =>
((-7) - 3) = ((3 - ((-7) / x)) - 3) | sub 3 =>
((-7) - 3) = ((3 - 3) - ((-7) / x)) | sub_comm 4, ((3 - ((-7) / x)) - 3) =>
((-7) - 3) = (0 - ((-7) / x)) | eval 5, (3 - 3) =>
(-10) = (0 - ((-7) / x)) | eval 1, ((-7) - 3) =>
-10x = ((0 - ((-7) / x)) * x) | mul x =>
(-10x / (-10)) = (((0 - ((-7) / x)) * x) / (-10)) | div (-10) =>
((x * (-10)) / (-10)) = (((0 - ((-7) / x)) * x) / (-10)) | comm 2, -10x =>
(x * ((-10) / (-10))) = (((0 - ((-7) / x)) * x) / (-10)) | assoc 1, ((x * (-10)) / (-10)) =>
(x * 1) = (((0 - ((-7) / x)) * x) / (-10)) | eval 3, ((-10) / (-10)) =>
x = (((0 - ((-7) / x)) * x) / (-10)) | mul1 1, (x * 1) =>
x = ((0x - (((-7) / x) * x)) / (-10)) | dist 3, ((0 - ((-7) / x)) * x) =>
x = ((0x - (x * ((-7) / x))) / (-10)) | comm 7, (((-7) / x) * x) =>
x = ((0x - ((x * (-7)) / x)) / (-10)) | assoc 7, (x * ((-7) / x)) =>
x = ((0x - (-7x / x)) / (-10)) | comm 8, (x * (-7)) =>
x = ((0 - (-7x / x)) / (-10)) | mul0 4, 0x =>
x = ((0 - ((-7) * (x / x))) / (-10)) | assoc 5, (-7x / x) =>
x = ((0 - ((-7) * 1)) / (-10)) | div_self 7, (x / x) =>
x = ((0 - (-7)) / (-10)) | eval 5, ((-7) * 1) =>
x = (7 / (-10)) | eval 3, (0 - (-7)) =>
x = ([-7/10]) | eval 2, (7 / (-10))
\end{verbatim}

\begin{verbatim}
(2 + 8x) = (-2x + 10) =>
((2 + 8x) - -2x) = ((-2x + 10) - -2x) | sub -2x =>
((2 + 8x) - -2x) = ((10 + -2x) - -2x) | comm 11, (-2x + 10) =>
((2 + 8x) - -2x) = (10 + (-2x - -2x)) | assoc 10, ((10 + -2x) - -2x) =>
((2 + 8x) - -2x) = (10 + 0) | sub_self 12, (-2x - -2x) =>
(2 + (8x - -2x)) = (10 + 0) | assoc 1, ((2 + 8x) - -2x) =>
(2 + ((8 - (-2)) * x)) = (10 + 0) | dist 3, (8x - -2x) =>
(2 + 10x) = (10 + 0) | eval 4, (8 - (-2)) =>
(10x + 2) = (10 + 0) | comm 1, (2 + 10x) =>
((10x + 2) - 2) = ((10 + 0) - 2) | sub 2 =>
(10x + (2 - 2)) = ((10 + 0) - 2) | assoc 1, ((10x + 2) - 2) =>
(10x + 0) = ((10 + 0) - 2) | eval 5, (2 - 2) =>
10x = ((10 + 0) - 2) | add0 1, (10x + 0) =>
(10x / 10) = (((10 + 0) - 2) / 10) | div 10 =>
((x * 10) / 10) = (((10 + 0) - 2) / 10) | comm 2, 10x =>
(x * (10 / 10)) = (((10 + 0) - 2) / 10) | assoc 1, ((x * 10) / 10) =>
(x * 1) = (((10 + 0) - 2) / 10) | eval 3, (10 / 10) =>
x = (((10 + 0) - 2) / 10) | mul1 1, (x * 1) =>
x = ((10 - 2) / 10) | eval 4, (10 + 0) =>
x = (8 / 10) | eval 3, (10 - 2) =>
x = [4/5] | eval 2, (8 / 10)
\end{verbatim}
\end{footnotesize}

\subsection{\texttt{fractions}}

The \texttt{fractions} environment exercises the ability to
reason about integer factorizations, especially common divisors and common multiples,
from primitive axioms.
A state in this environment is one of:

\begin{description}
    \item[Number: ]An integer $n$,
    \item[Number operations: ]Either addition, subtraction or multiplication of two terms, both of which can be either numbers or number operations,
    \item[Fraction: ]A single fraction, where numerator and denominator are either numbers or number operations,
    \item[Fraction operation: ] An operation ($+$, $-$ or $\times$) between two fractions, two numbers, or a fraction and a number.
\end{description}

A state is solved if it is either a number or a fraction where both
numerator and denominators are numbers that are coprime
(i.e. their greatest common divisor has to be $1$).
Note that a fraction operation can only involve two fractions, not other
(recursively defined) fraction operations.
This is to keep this domain testing an orthogonal skill compared to \texttt{equations}:
nested operations would require more elaborate algebraic manipulations,
but this environment focuses on the Common Core topic of fraction manipulation.

\begin{table}
    \centering
    \caption{Axioms of the \texttt{fractions} domain.}
    \label{tab:fr-actions}
    \begin{tabular}{p{1.7cm} p{7cm} p{4cm}}
        \toprule
        Mnemonic & Description & Example \\
        \midrule
        factorize & Factorize a composite integer into a prime factor times a divisor. & $\frac{20}{5} \rightarrow \frac{5\times 4}{5}$ \\
        & & \\
        
        cancel & Eliminate a common factor between both the numerator and the denominator. Only applies when the factor is explicitly written in both expressions. & $\frac{2 \times 5}{5 \times 10} \rightarrow \frac{2}{10}$ \\
        & & \\
        
        eval & Evaluate an operation with numbers. & $\frac{2 \times 5}{10} \rightarrow \frac{10}{10}$ \\
        & & \\

        scale & Multiply both the numerator and denominator of a fraction by a prime $p \in \left\{2, 3, 5, 7\right\}$. & $\frac{1}{2} + \frac{1}{6} \rightarrow \frac{1\times3}{2\times3} + \frac{1}{6}$ \\
        & & \\
    
        simpl1 & Replace a fraction with denominator $1$ by its numerator. & $\frac{10 + 5}{1} \rightarrow 10 + 5$ \\
        & & \\
        
        mfrac & Rewrite a number as a fraction with denominator $1$. & $5 + \frac{2}{3} \rightarrow \frac{5}{1} + \frac{2}{3}$ \\
        & & \\
        
        mul & Multiply two fractions. & $\frac{3}{4} \times \frac{2}{3} \rightarrow \frac{3\times2}{4\times3}$ \\
        & & \\
        
        combine & Add or subtract two fractions that have syntactically equal denominators. & $\frac{3}{4 + 1} + \frac{9\times2}{4 + 1} \rightarrow \frac{3 + (9\times2)}{4 + 1}$ \\

        \bottomrule
    \end{tabular}
\end{table}

The following are three random problems solved by ConPoLe demonstrating all axioms:

\begin{footnotesize}

\begin{verbatim}
[1]/[105] + [1]/[42] =>
[1]/[105] + [(5 * 1)]/[(5 * 42)] | scale 4, 5 =>
[1]/[105] + [(5 * 1)]/[210] | eval 8, 5 * 42 =>
[(2 * 1)]/[(2 * 105)] + [(5 * 1)]/[210] | scale 1, 2 =>
[(2 * 1)]/[210] + [(5 * 1)]/[210] | eval 5, 2 * 105 =>
[((2 * 1) + (5 * 1))]/[210] | combine 0 =>
[(2 + (5 * 1))]/[210] | eval 2, 2 * 1 =>
[(2 + 5)]/[210] | eval 3, 5 * 1 =>
[7]/[210] | eval 1, 2 + 5 =>
[7]/[(7 * 30)] | factorize 2, 210, 7*30 =>
[1]/[30] | cancel 0, 7
\end{verbatim}

\begin{verbatim}
[18]/[5] - 1 =>
[18]/[5] - [1]/[1] | mfrac 4, 1 =>
[18]/[5] - [(5 * 1)]/[(5 * 1)] | scale 4, 5 =>
[18]/[5] - [5]/[5] | cancel 4, 1 =>
[(18 - 5)]/[5] | combine 0 =>
[13]/[5] | eval 1, 18 - 5
\end{verbatim}

\begin{verbatim}
5 * 3 =>
[5]/[1] * 3 | mfrac 1, 5 =>
[5]/[1] * [3]/[1] | mfrac 4, 3 =>
[(5 * 3)]/[(1 * 1)] | mul 0 =>
[(5 * 3)]/[1] | eval 4, 1 * 1 =>
[15]/[1] | eval 1, 5 * 3 =>
15 | simpl1 0
\end{verbatim}

\end{footnotesize}

We used a custom generator for fraction problems. First, with 25\% chance,
we choose to generate a single-term problem; otherwise, we will generate
a \emph{fraction operation} (two terms with an operator drawn uniformly from $\{+, -, \times\})$. We then generate the subterms independently as follows. With $50\%$ chance, we generate
a number. A number is generated by first picking the number of prime
factors (between 0 and 4), then drawing each factor independently
from the set $\{2, 3, 5, 7\}$ and multiplying them.
A fraction is generated by generating two numbers with the same described
procedure: the first becomes the numerator, and the second becomes the denominator.

\subsection{\texttt{ternary-addition}}

The \texttt{ternary-addition} domain exercises step-by-step arithmetic, in an
analogous fashion to some example-tracing arithmetic tutors \cite{weitekamp2020interaction},
where operations can be performed out of the traditional order as long as they are correct deductions.
Each state is a sequence of digits multiplying powers of $3$, that are being added
together. Two digits can be combined (added together) when they are adjacent and multiply the
same power (e.g. $2\times3^3$ and $1\times3^3$ can be combined together, but $2\times3^3$ and
$1\times3^5$ cannot). Three operations are available: (a) combining two adjacent digits
that multiply the same power -- generating two other digits, (b) swapping any pair of adjacent
digits, and (c) deleting a digit $0$ from anywhere. A state is solved when the final
number can be readily read from the state: all digits must multiply different powers,
they must be sorted by power, and there should be no zero digits.
For example, $2\times3^3 + 1\times3^5$ is simplified.
On the other hand, $2\times3^3 + 1\times3^5 + 1\times3^3$ is not: the digits multiplying
$3^3$ can be brought together and further combined.

To represent digits and powers as strings, we use the letters $a, b, c$ to represent digits $0, 1, 2$ respectively, and decimal digits $0-9$ to represent powers.
There is an implicit addition operation between all digits in the state.
For example, $c3~b5~b3$ represents $2\times3^3 + 1\times3^5 + 1\times3^3$.
Table~\ref{tab:td-actions} lists the three axioms described above, with examples.

\begin{table}
    \centering
    \caption{Axioms of the \texttt{ternary-addition} domain.}
    \label{tab:td-actions}
    \begin{tabular}{p{1.7cm} p{7cm} p{4cm}}
        \toprule
        Mnemonic & Description & Example \\
        \midrule
        swap & Swap any two adjacent digits & $b3~b5~c3 \rightarrow b3~c3~b5$ \\
        & & \\
        
        comb & Combine (add) two adjacent digits that multiply the same power $p$, replacing them by two other digits: the result (which has power $p$) and the carry (with power $p+1$). & $b3~c3~b5 \rightarrow a3~b4~b5$ \\
        & & \\
        
        del & Erase a digit $0$ ($a$). & $a3~b4~b5 \rightarrow b4~b5$ \\
        \bottomrule
    \end{tabular}
\end{table}

The following are two of ConPoLe's solutions for random problems, both utilizing all 3 axioms.

\begin{footnotesize}
\begin{verbatim}
#(c3 c3 b5 b5 b5 a1 a0 c0) =>
#(c3 c3 b5 c5 a6 a1 a0 c0) | comb 3, b5 b5 =>
#(c3 c3 a5 b6 a6 a1 a0 c0) | comb 2, b5 c5 =>
#(c3 c3 a5 b6 a6 a1 c0) | del 6, a0 =>
#(c3 c3 a5 b6 a6 c0) | del 5, a1 =>
#(c3 c3 a5 b6 c0) | del 4, a6 =>
#(c3 c3 a5 c0 b6) | swap 3, b6 c0 =>
#(c3 c3 c0 b6) | del 2, a5 =>
#(c3 c0 c3 b6) | swap 1, c3 c0 =>
#(c0 c3 c3 b6) | swap 0, c3 c0 =>
#(c0 b3 b4 b6) | comb 1, c3 c3
\end{verbatim}

\begin{verbatim}
#(a1 b5 c1 b3 c3 b5 a2 c1 c1 c1 b0 b3 a5 b5) =>
#(a1 b5 c1 b3 c3 b5 a2 c1 b1 b2 b0 b3 a5 b5) | comb 8, c1 c1 =>
#(a1 b5 c1 b3 c3 b5 a2 a1 b2 b2 b0 b3 a5 b5) | comb 7, c1 b1 =>
#(b5 c1 b3 c3 b5 a2 a1 b2 b2 b0 b3 a5 b5) | del 0, a1 =>
#(b5 c1 b3 c3 b5 a2 b2 b2 b0 b3 a5 b5) | del 6, a1 =>
#(b5 c1 a3 b4 b5 a2 b2 b2 b0 b3 a5 b5) | comb 2, b3 c3 =>
#(b5 c1 b4 b5 a2 b2 b2 b0 b3 a5 b5) | del 2, a3 =>
#(c1 b5 b4 b5 a2 b2 b2 b0 b3 a5 b5) | swap 0, b5 c1 =>
#(c1 b5 b4 b5 b2 b2 b0 b3 a5 b5) | del 4, a2 =>
#(c1 b5 b4 b5 c2 a3 b0 b3 a5 b5) | comb 4, b2 b2 =>
#(c1 b4 b5 b5 c2 a3 b0 b3 a5 b5) | swap 1, b5 b4 =>
#(c1 b4 c5 a6 c2 a3 b0 b3 a5 b5) | comb 2, b5 b5 =>
#(c1 b4 c5 c2 a3 b0 b3 a5 b5) | del 3, a6 =>
#(c1 b4 c5 c2 b0 b3 a5 b5) | del 4, a3 =>
#(c1 b4 c5 b0 c2 b3 a5 b5) | swap 3, c2 b0 =>
#(c1 b4 b0 c5 c2 b3 a5 b5) | swap 2, c5 b0 =>
#(c1 b0 b4 c5 c2 b3 a5 b5) | swap 1, b4 b0 =>
#(b0 c1 b4 c5 c2 b3 a5 b5) | swap 0, c1 b0 =>
#(b0 c1 b4 c5 c2 b3 b5) | del 6, a5 =>
#(b0 c1 b4 c2 c5 b3 b5) | swap 3, c5 c2 =>
#(b0 c1 b4 c2 b3 c5 b5) | swap 4, c5 b3 =>
#(b0 c1 b4 c2 b3 a5 b6) | comb 5, c5 b5 =>
#(b0 c1 b4 c2 b3 b6) | del 5, a5 =>
#(b0 c1 c2 b4 b3 b6) | swap 2, b4 c2 =>
#(b0 c1 c2 b3 b4 b6) | swap 3, b4 b3
\end{verbatim}
\end{footnotesize}

To generate a problem, we first pick the number of digits in the sequence uniformly from $1$ to $15$. Then, we choose each element independently, by choosing a digit from $\{0, 1, 2\}$ and
a power from $\{0, 1, 2, 3, 4, 5, 6\}$, all independently and uniformly.

\subsection{\texttt{sorting}}

The \texttt{sorting} environment tests the ability to measure and compare object lengths,
inspired by the ``Measurements and Data'' section from Common Core.
States in this domain are a permutation of the integers from $1$ to $L$, where $L$
is the length of the list. When represented as a string, each number $n_i$
is written as a repetition of the \texttt{=} character, $n_i$ times; \texttt{|} is used
as a separator between numbers.
The goal is to sort the list by the length of each of the substrings.
Table~\ref{tab:sr-actions} lists the only two axioms in this domain:
swapping adjacent elements and reversing the list.
Below, we show two solutions generated by ConPoLe. The first
is done with swaps only. In the second problem, the reversed list
has less inversions than the given one: ConPoLe learns to first reverse
the list, and then sort the result using swaps.

\begin{table}
    \centering
    \caption{Axioms of the \texttt{sorting} domain.}
    \label{tab:sr-actions}
    \begin{tabular}{p{1.7cm} p{4cm} p{7cm}}
        \toprule
        Mnemonic & Description & Example \\
        \midrule
        swap & Swap two adjacent elements. & \texttt{[=|==|====|===]} $\rightarrow$ \texttt{[=|==|===|====]} \\
        reverse & Reverse the entire list. & \texttt{[===|==|=]} $\rightarrow$ \texttt{[=|==|===]} \\
        \bottomrule
    \end{tabular}
\end{table}

\begin{footnotesize}
\begin{verbatim}
[====|==|=|===|=====|======] =>
[====|=|==|===|=====|======] | swap 1 =>
[=|====|==|===|=====|======] | swap 0 =>
[=|==|====|===|=====|======] | swap 1 =>
[=|==|===|====|=====|======] | swap 2
\end{verbatim}

\begin{verbatim}
[========|======|===|=|==|====|=======|=====]
[=====|=======|====|==|=|===|======|========] | reverse =>
[=====|====|=======|==|=|===|======|========] | swap 1 =>
[=====|====|==|=======|=|===|======|========] | swap 2 =>
[=====|====|==|=|=======|===|======|========] | swap 3 =>
[=====|====|==|=|===|=======|======|========] | swap 4 =>
[=====|==|====|=|===|=======|======|========] | swap 1 =>
[=====|==|====|=|===|======|=======|========] | swap 5 =>
[=====|==|=|====|===|======|=======|========] | swap 2 =>
[=====|==|=|===|====|======|=======|========] | swap 3 =>
[=====|=|==|===|====|======|=======|========] | swap 1 =>
[=|=====|==|===|====|======|=======|========] | swap 0 =>
[=|==|=====|===|====|======|=======|========] | swap 1 =>
[=|==|===|=====|====|======|=======|========] | swap 2 =>
[=|==|===|====|=====|======|=======|========] | swap 3
\end{verbatim}
\end{footnotesize}

For generating problems, we first choose a length $L$ uniformly from $2$ to $11$, and then
shuffle the list of integers from $1$ to $L$.
Lists of $11$ elements have at most $55$ inversions. Therefore, because of the \texttt{reverse} operation, all of them can be sorted with at most $27$ adjacent swaps (plus one use of \texttt{reverse}, potentially).

\section{Training and architecture details}

\textbf{Training/test split}. The generators described in Appendix~\ref{sec:ap-commoncore} use pseudo-random number generators,
and thus are deterministic if the random seed is fixed. We use this fact to generate distinct
training and test environments. For training, agents start with a random seed given by
a OS-provided randomness source. Every time an agent samples a new problem, a new seed
is chosen from $10^6$ to $10^7$ (providing around $10^7$ potential training problems).
For testing, we always use the seeds from $0$ to $199$, providing $200$ training problems.

\textbf{Architecture details}. All models use character-level bidirectional LSTM encoders.
We first use $64$-dimensional character embeddings. Then, we use two stacked bi-LSTM layers,
with a hidden dimension of $256$. Finally, we take the last hidden state of each direction
from the last layer, concatenate their vectors to obtain a $512$-dimensional embedding
of the state, and transform this embedding with a $2$-layer MLP that preserves dimension
(and do the same separately for the action, in DRRN), and use that final output
according to each model's architecture.
ConPoLe learns a $512 x 512$ matrix $W_\theta$ that performs
the bilinear transform; CVI learns a linear layer, and DRRN embeds state and action and outputs
their dot product.

\textbf{Hyper-parameters}. We first picked the learning rate from $10^{-i}$ and $5\times10^{-i}$ for $i$ ranging from $1$ to $6$; in shorter experiments of 100k environment steps in \texttt{equations} and \texttt{fractions}, the value of $5\times10^{-6}$ had the highest success rate for CVI and ConPoLe (though the difference to $10^{-5}$ and $5\times10^{-5}$ was insignificant); for DRRN, $10^{-4}$ performed best on average.
We thus used these values in all experiments.
Next, we picked the frequency of updates and batch sizes.
For ConPoLe and CVI, we observed that more frequent updates were consistently better;
for performance, we chose to optimize every $10$ solved problems, taking $256$ gradient steps
on randomly sampled contrastive examples from the replay buffer.
For DRRN, since each training example requires computing a $\max$ operation
for the $Q$ update, we chose a smaller batch size to keep training runs in a single domain
under $3$ days. We therefore picked a batch size of $64$, and performed training
updates every $16$ problems.

\end{document}